\documentclass[runningheads]{llncs}
\usepackage{graphicx}
\usepackage{wrapfig}
\usepackage{subfigure}
\usepackage{hyperref}

\usepackage{xcolor}
\usepackage{orcidlink}

\begin{document}

\title{Data augmentation on graphs for table type classification}
\titlerunning{Data augmentation on graphs for table type classification}
%
\author{Davide del Bimbo\orcidlink{0000-0002-2834-1283}  \and
Andrea Gemelli\orcidlink{0000-0002-6149-8282}
\and Simone Marinai\orcidlink{0000-0002-6702-2277}}
\authorrunning{D. del Bimbo et al.}
%
\institute{University of Florence, Florence, Italy \\ AI Lab, DINFO \\ \email{davide.delbimbo@stud.unifi.it} \\ \email{\{andrea.gemelli, simone.marinai\}@unifi.it}}
\maketitle              

\begin{abstract}
Tables are widely used in documents because of their compact and structured representation of information.
In particular, in scientific papers, tables can sum up novel discoveries and summarize experimental results, making the research comparable and easily understandable by scholars.
Since the layout of tables is highly variable, it would be useful to interpret their content and classify them into categories. This could be helpful to directly extract information from scientific papers, for instance comparing performance of some models given their paper result tables. 
In this work, we address the classification of tables using a Graph Neural Network, exploiting the table structure for the message passing algorithm in use. We evaluate our model on a subset of the Tab2Know dataset. Since it contains  few examples manually annotated, we propose data augmentation techniques directly on the table graph structures.
We achieve promising preliminary results, proposing a data augmentation method suitable for graph-based table representation.

\end{abstract}

\section{Introduction}\label{ch:introduzione}
\label{ch:introduction}

Tables within scientific documents represent an essential source of knowledge. Their use is necessary for the intelligibility of a document as they provide useful information in a structured and well-organized form, allowing the reader to understand the data through visual content.
In particular, in scientific documents, the tables can summarize data from experiments, observations and much more, providing essential information to reconstruct the state of the art of different fields of research \cite{kardas2020axcell}.
Since different users write different documents, tables usually present different layouts: sometimes, they can be irregular or contain unique abbreviations that are difficult to disambiguate automatically. 
It would be helpful if their contents were interpreted and transcribed into a Knowledge Base (KB), a database in which tables are translated using a single standard vocabulary. The use of the KB could be helpful to those who need to make use of the information and data contained in the tables without having to access the documents directly \cite{Kruit2020Tab2KnowBA}.
In this scenario, it appears necessary to define a way to classify tables into entities that share common features.
 
In this work we present a model to classify scientific tables given their content and structure. The label of a table is related to its purpose within the paper and, as proposed in \ \cite{Kruit2020Tab2KnowBA}, we try to classify them into four different types: \emph{Observation}, \emph{Input}, \emph{Example} and \emph{Other}.
This classification is useful in areas such as the automatic comprehension of an article or the summarization of information in a document.
To address the task just described, we make use of Graph Neural Networks (GNNs), which have been widely considered recently in Document Analysis and Table Understanding. This choice is motivated by their ability to consider the structural information. In addition, we propose some data augmentation techniques working directly on the graph representation of tables, which led to promising preliminary results.

This  work is organized as follows.  In Section\ \ref{ch:related-works} works that mostly inspired our paper are explored, focusing on the most significant ones. The proposed approach\footnote{Code available \href{https://github.com/AILab-UniFI/DA-GraphTab}{\underline{here}}} is discussed in Section\ \ref{se:method} including the preprocessing of the tables of scientific papers, the  data augmentation techniques and the 
implementation of the GNN model. Experimental results on the Tab2Know dataset are presented in Section\ \ref{ch:chapter4}, while conclusions are drawn in Section\ \ref{sec:concl}.

\section{Related Works}
\label{ch:related-works}

In this section we summarize previous work related to the proposed approach.

\textbf{Table related tasks.}
Usually, to extract information from tables in documents two  steps are used: first tables are detected, then their structure is described in terms of rows and columns. 
As shown in \cite{hashmi2021current} different techniques have been used in the past to tackle these tasks, making use of both computer vision and natural language processing techniques. 
Recently, two new approaches have beaten the state-of-the-art: combining vision, semantic and relations for layout analysis and table detection \cite{zhang2021vsr} and applying a soft pyramid mask learning mechanism in both the local and global feature maps for complicated table structure recognition \cite{qiao2021lgpma}.
In addition to Table Detection (TD) and Table Structure Recognition (TSS), the authors who released PubTables-1M \cite{DBLP:journals/corr/abs-2110-00061} proposed to perform Functional Analysis (FA) to distinguish table headers from table cells. 
In \cite{gemelli2022} we proposed a Graph Neural Network method to perform FA along with TD, TSS and document layout analysis to enrich the information of extracted tables with a context.

\textbf{Information extraction from scientific literature.}
Automatic extraction of table information can help scholars in several disciplines. In addition to values in table cells it is also useful to classify tables according to their type.
To track progresses in scientific research, authors of \cite{kardas2020axcell} propose an automatic machine learning pipeline for extracting results from papers. The pipeline is split into three steps, the first one being table type classification. Since the focus is on results extraction, result and ablation tables are identified. The extracted information is summarized into a leaderboard, sorted by the best scores given certain metrics.
Another work, Tab2Know \cite{Kruit2020Tab2KnowBA}, proposes to classify tables in four types and recognize table headers and columns. 
The aim is to extract and link tables into a knowledge base to answer user queries trying to identify relevant information over years of research in a given field.
Table classification is referred by the authors as  table type detection.

\textbf{Data Augmentation techniques.} The Tab2Know dataset can be used for performing table classification. However, the manually labeled subset is small and therefore we need to implement suitable Data Augmentation (DA) techniques. DA is widely used in machine learning in order to make models better generalize on unseen samples and unbalanced datasets.
In object detection, DA techniques involve color operations (contrast, brightness), geometric operations (translations, rotations), and bounding box operations \cite{zoph2020learning}. 
None of these can be used in our case since we are considering graphs to represent the tables and augmentation operations commonly used in vision and language have no analogs for graphs \cite{DBLP:journals/corr/abs-2006-06830}.
Similarly to what we did for trees \cite{baldi2003using} and inspired by \cite{khan2021tabaug}, we applied some of their augmentations on table examples directly in the graph structure (Section \ref{sub:data-augmentation}). 
Operations that can be performed on tables are random deletion of rows, row replication, column deletion and column replication. Instead of working directly on images, we therefore extract the table structure (Section \ref{sub:preprocessing}) and then apply DA on their graph representation, by means of node deletion, edge deletion and inversion of node contents (Section \ref{sub:data-augmentation}).
\section{Method}
\label{se:method}

In this section we present the main steps of the proposed approach.

\subsection{Preprocessing}
\label{sub:preprocessing}

\begin{figure}[t]
    \centering
    \includegraphics[width=0.6\textwidth]{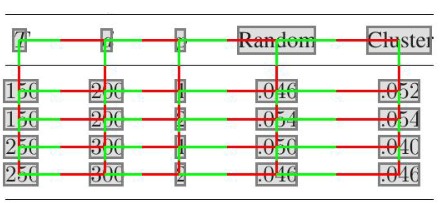}
    \caption{Words and bounding boxes  extracted from one PDF paper using PyMuPDF. Nodes are connected through a visibility graph.}
    \label{fig:graphTable}
\end{figure}

The first step to apply a GNN for table classification is the conversion of tables in PDF papers in graphs.
To this purpose, we use PyMuPDF, a toolkit for viewing and rendering PDF and XPS files \cite{McKie2022pymupdf}.
The library is used to extract words and their bounding boxes; by using the positions of the tables in the annotations, only the words within them can be considered (Fig. \ref{fig:graphTable}). One graph for each table is built, where words correspond to nodes.
A feature vector is associated to each word and contains information about its position and the embedding of the textual content, extracted using spaCy language models \cite{honnibal2017natural}.
Edges represent the mutual position of bounding boxes and are identified by a visibility graph, like the one described in \cite{riba2019table} (e.g. see Fig. \ref{fig:graphTable}). 
Each node is connected to its nearest visible nodes when their bounding boxes intersect horizontally or vertically.
Each graph, representing a table, is associated with the annotation corresponding to its type.

\begin{figure}[!h]
    \centering
    \includegraphics[width=0.7\textwidth]{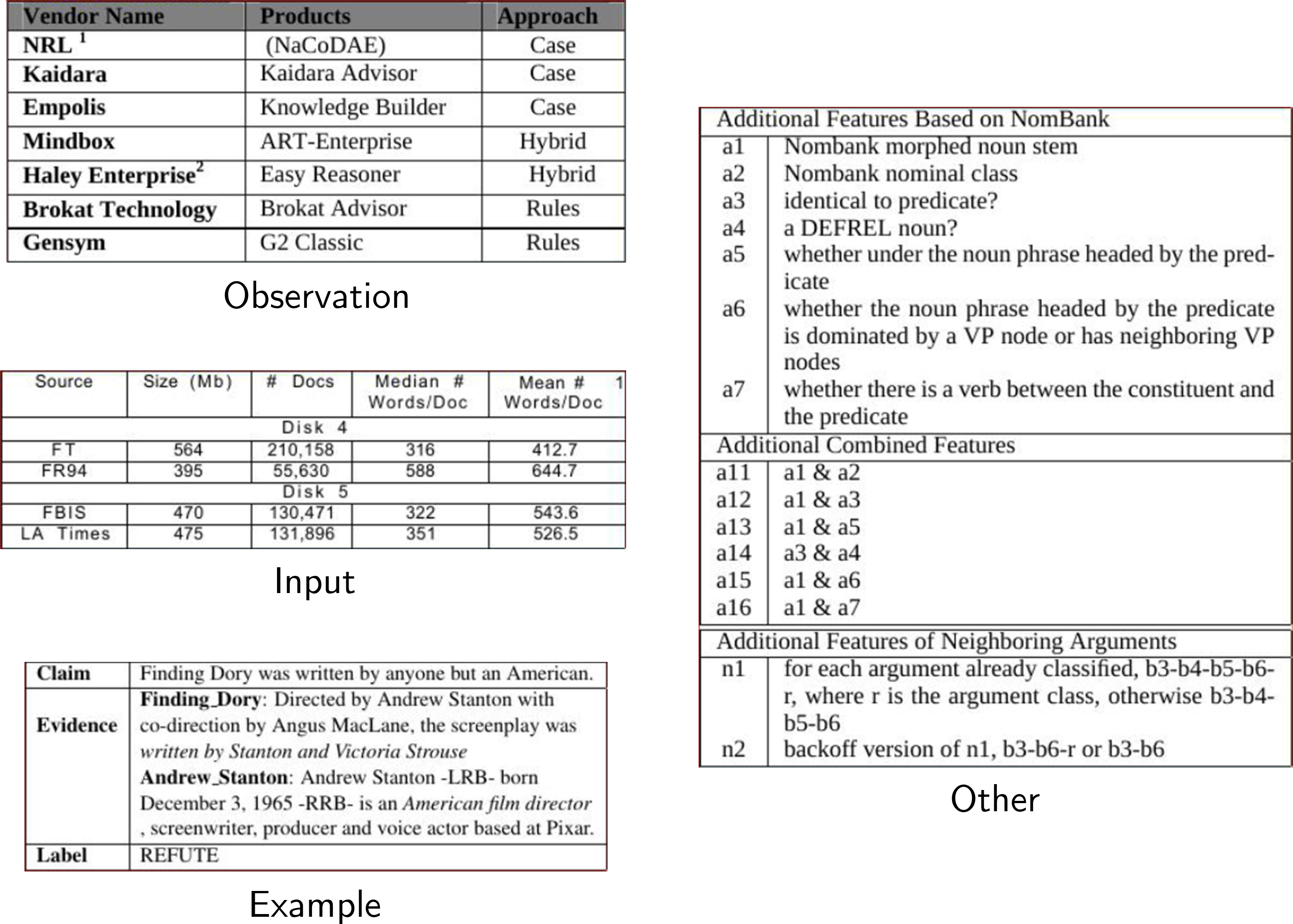}
    \caption{Different types of tables with their classes.}
    \label{fig:tableClass}
\end{figure}

Tables without annotation and those of which a graph cannot be built are discarded. At the end we obtain 320 graphs split into four classes:
Observation (235), Input (43), Example (13), and Other (29). 
Examples of classes can be seen in Fig. \ref{fig:tableClass}
Each node in the graph corresponds to a feature vector. In addition to the geometric features of the nodes, such as position and size,
textual content embeddings are added using spaCy.
In particular, two spaCy models are used and compared: {\em en\_core\_web\_lg} and {\em en\_core\_sci\_lg}. 
The first one is the largest english vocabulary which associates each word with a numerical vector of 300 values; the other model, trained on a biomedical corpus, associates each word a numerical vector of 200 values. 
The results obtained using the two models are compared in the experiments.

\subsection{Data Augmentation}
\label{sub:data-augmentation}

\begin{figure}[t]
        \centering
        \includegraphics[width=0.66\textwidth]{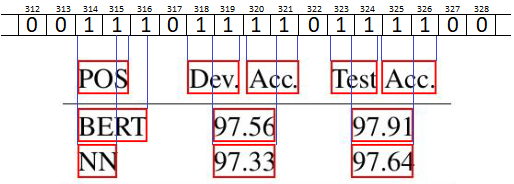}
        \caption{Recognition of columns. Group of 1s in the projected vector indicate different columns.}
        \label{fig:columnRec}
    \end{figure}

\begin{figure}[t]
        \centering
        \includegraphics[width=0.5\textwidth]{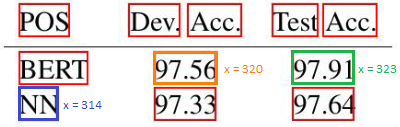}
        \caption{Recognition of rows. The blue bounding box is detected belonging to a new row since its \textit{x} coordinate is lower than the previous green block.}
        \label{fig:rowRec}
    \end{figure}

Since the dataset (Section\ \ref{se:dataset})  is unbalanced, a Data Augmentation strategy  has been implemented to generate  new training data from the available ones.
For DA,  new graphs can be obtained by modifying their structure and the information associated with the nodes and edges.
Since the embedding for each table is evaluated through a message passing algorithm that strongly relies on the table structure and content, removing elements of the graph and changing node features helps to generate more variability of examples for each class. This not only improves the generalization capability of the model, but can help to  reduce the class imbalance.

\textbf{Random removal of nodes and edges}
In these operations, a random sample of nodes or arcs within the table is removed from the graph.
By doing so, it is possible to generate a new graph similar to the initial one, but with different information.
\begin{itemize}
    \item nodes removal: a random subset of node indexes is removed. The size of the sample depends on the number of nodes in the graph, a random number between 1\% and 20\% of the total number of nodes.
    
    \item edges removal: a random subset of edge indexes is removed. The size of the sample depends on the number of edges in the graph, a random number between 1\% and 20\% of the total number of edges.
\end{itemize}
The amount of randomly removed nodes/edges is an arbitrary choice. We did not want to: (i) discard too much information and (ii) introduce any bias in the decision.

\textbf{Inversion of rows and columns}
The row and column inversion technique is more complex, due to the fact that the internal structure of the tables is not known. Therefore, it is necessary to define an approach to approximate this structure. Once identified, rows or columns can be inverted, by means of swapping their node features.
\begin{itemize}
    \item Column inversion:
    table columns identification is made with a projection-profile based approach which
    defines a vector of size equal to the width of the table region. Each element of the vector is initialized to 0. Then, for each word, the coordinates $x_1$ and $x_2$ of the corresponding bounding box are extracted and projected, setting to 1 the vector values whose indices correspond to these coordinates. The obtained result is shown in figure \ref{fig:columnRec}: adjacent 0s should identify column boundaries, while adjacent 1s the coordinates of each column. Thus, two columns can be inverted by swapping their contents, that is, the features of the nodes whose center of the bounding box belongs to those columns.
    The limitation of this technique is visible whenever there is a space between words belonging to the same column.
    \item Rows inversion: To reverse rows, it is necessary to compare the positions of "successive" bounding boxes. PyMuPDF reads and orders the content from left to right and from top to bottom. So, when a bounding box appears positioned ahead of the next one, it means that the latter is on a new row. In Fig. \ref{fig:rowRec} the orange, green and blue bounding boxes are successive ones: the last one is on a new row since its \textit{x} coordinate is lower than the green one. Once the structure of the rows has been identified, they can be reversed by swapping the features of the nodes belonging to them. 
    The limitation of this technique is visible in the case of multi-row tables.
\end{itemize}

\subsection{Model}
\label{se:model}
Our baseline model uses two Graph Convolutional layers \cite{DBLP:journals/corr/KipfW16} and a Linear output one.
Each node embedding is updated through the message passing algorithm: (i) firstly each node collects the embeddings of connected nodes; in our case study, each cell of the tables collects the embeddings of the visible ones; (ii) then a weighted sum is applied to aggregate the collected information and to update each node vector. (iii) At the end, a fully connected layer is used to learn the new node representations between the layers of the network
We applied this procedure twice and then all the nodes of each graph are aggregated using a \textit{redout} function. 
Every graph in the data may have its unique structure, as well as its node and edge features. In order to make a single prediction per each table, we aggregated and summarized over all the node information.
Given a graph, the average node feature readout we use is 
\[h_g = \frac{1}{|V|}\sum_{v\in V}h_v\]
where $h_g$ is the representation of graph $g$, $V$ is the set of nodes in $g$, $h_n$ is the feature of node $n$.
\section{Experiments}
\label{ch:chapter4}

In this section, we present the dataset used in the experiments that are subsequently discussed.

\subsection{The Tab2Know dataset}
\label{se:dataset}
The Tab2Know dataset, proposed in \cite{Kruit2020Tab2KnowBA}, contains information regarding tables extracted from scientific papers in the Semantic Scholar Open Research Corpus. Tables are extracted using PDFFigures \cite{pdffigures2}, a tool that finds figures, tables, and captions within PDF documents, and Tabula\footnote{\href{https://github.com/tabulapdf/tabula}{https://github.com/tabulapdf/tabula}}, that outputs a CSV per each table reflecting its structure and content. After the conversion, each table is saved as an RDF triple addressable by an unique URI. Each CSV is then analyzed to recognize headers, type of table and columns type. 
The authors define an ontology of 27 different classes, 4 of which are defined as "root" ones (Example, Input, Observation and Other): the others are given depending on the type of columns found inside each table (e.g. Recall is a subclass of Metric that is a subclass of Observation). Their training corpus is composed of 73k tables, labeled using Snorkel \cite{ratner2017snorkel} and starting from a small pre-labeled set of tables obtained through human supervision using SPARQL queries. Human annotators then looked at 400 of them, checking their labeling correctness and, after resolving their conflicts when disagreeing, used this subset as the test set.

\subsection{Using the dataset}

To extract and group information on tables from Tab2Know, we built a conversion system to derive a JSON object for each available table. 
The information is the table numbering in the document, the page number where the table is located, the number of rows that make up the header, the document URL, the table class definition, and the caption text. We also added some information not represented in the RDF graph, such as the position of the table and the location of the caption (the latter information is  obtained using PDFFigures and Tabula). Then we downloaded tables corresponding PDF papers, accessed from the Semantic Scholar Open Research Corpus. From each paper, the pages containing the tables are extracted.
Unfortunately it is not possible to  use the whole Tab2Know dataset. For instance, some papers are no longer available or an updated version do not match anymore the annotations provided. From the total, only the data whose annotations match are used, discarding the others. We obtained a subset containing 33,069 tables extracted from 11,800 scientific documents (45\% of the original one). In addition, this dataset is very unbalanced (80\% Observation, 10\% Input, 7\% Other, 3\% Example) and it contains several missing or wrong annotations (55\% of column classes have been labeled as 'others', across 22 different classes).
For these reasons, we only use in this preliminary work the test set that was manually classified and corrected by humans. Specifically, this dataset contains 361 tables extracted from 253 scientific papers. The distribution of tables according to the class is as follows: 235 \(Observation\), 43 \(Input\), 13 \(Example\), 29 \(Other\) (41 were 'unclassified', and we do not consider them during training).
We retain 20\% of this subset as training (randomly sampled keeping the same class occurrences) and, through the data augmentation techniques described before, we evaluated the generalization capabilities of the proposed model.

\subsection{Results}

\begin{table}[t]
    \caption{\label{tab:allRes}Results without data augmentation (\emph{No Aug.}); Data Augmentation with Rows and Columns inversion (\emph{R/C}); Data Augmentation with Rows and Columns inversion and random removal of nodes and edges (\emph{All}).  $P$, $R$ and $F1$ correspond to Precision, Recall and F1 score.}
    \centering
    \fontsize{9}{10}\selectfont
    \begin{tabular}{|c||c|c|c||c|c|c||c|c|c||c|c|c|}
        \cline{2-7}
        \multicolumn{1}{c|}{} & \multicolumn{6}{c|}{\emph{No Aug.}} & \multicolumn{6}{c}{} \\ \cline{2-7} 
        \multicolumn{1}{c|}{} & \multicolumn{6}{c|}{train size: 63} & \multicolumn{6}{c}{} \\ \cline{2-7} 
        \multicolumn{1}{c|}{} & \multicolumn{3}{c|}{\textbf{web\_lg}} & \multicolumn{3}{c|}{\textbf{sci\_lg}} & \multicolumn{6}{c}{}\\
        \cline{1-7}
        \emph{Classes} (\#) & \emph{P} & \emph{R} & \emph{F1} & \emph{P} & \emph{R} & \emph{F1} & \multicolumn{6}{c}{}\\
        \cline{1-7}
        Observation (185) & 0.85 & 0.84 & 0.84 & 0.87 & 0.92 & 0.89\\
        \cline{1-7}
        Input (35) & 0.37 & 0.49 & 0.42 & 0.47 & 0.54 & 0.51\\
        \cline{1-7}
        Example (10) & 0.42 & 0.5 & 0.45 & 0.67 & 0.2 & 0.31\\
        \cline{1-7}
        Other (23) & 0.00 & 0.00 & 0.00 &0.43 & 0.26 & 0.32 \\
        \cline{1-7}\cline{1-7}
        \emph{All} (253) & 0.41 & 0.46 & 0.43 & 0.61 & 0.48 & \textbf{0.51} \\
        \cline{1-7}
        \multicolumn{13}{c}{}\\
        \cline{2-13}
        \multicolumn{1}{c|}{} & \multicolumn{12}{c|}{\emph{R/C}} \\ \cline{2-13}       
        \multicolumn{1}{c|}{} & \multicolumn{6}{c||}{train size: 200} & \multicolumn{6}{c|}{train size: 400} \\
        \cline{2-13}
        \multicolumn{1}{c|}{} & \multicolumn{3}{c|}{\textbf{web\_lg}} & \multicolumn{3}{c||}{\textbf{sci\_lg}} & \multicolumn{3}{c|}{\textbf{web\_lg}} & \multicolumn{3}{c|}{\textbf{sci\_lg}} \\
        \hline
        \emph{Classes} (\#) & \emph{P} & \emph{R} & \emph{F1} & \emph{P} & \emph{R} & \emph{F1} & \emph{P} & \emph{R} & \emph{F1} & \emph{P} & \emph{R} & \emph{F1} \\
        \hline
        Observation (185) & 0.82 & 0.84 & 0.83 & 0.85 & 0.92 & 0.89 & 0.82 & 0.84 & 0.84 & 0.84 & 0.94 & 0.88 \\
        \hline
        Input (35) & 0.37 & 0.43 & 0.39 & 0.52 & 0.46 & 0.48 & 0.34 & 0.34 & 0.34 & 0.52 & 0.40 & 0.45 \\
        \hline
        Example (10) & 0.80 & 0.40 & 0.53 & 0.60 & 0.30 & 0.40 & 0.57 & 0.40 & 0.47 & 0.50 & 0.30 & 0.37 \\
        \hline
        Other (23) & 0.06 & 0.04 & 0.05 & 0.41 & 0.30 & 0.35 & 0.05 & 0.04 & 0.05 &  0.50 & 0.30 & 0.38 \\
        \noalign{\hrule height 2pt}
        \emph{All} (253) & 0.51 & 0.43 & 0.45 & 0.60 & 0.50 & \textbf{0.53} & 0.45 & 0.41 & 0.42 & 0.59 & 0.48 & 0.52 \\   \hline
        \multicolumn{13}{c}{} \\ \cline{2-13}       
        \multicolumn{1}{c|}{} & \multicolumn{12}{c|}{\emph{All}} \\ \cline{2-13}       
        \multicolumn{1}{c|}{} & \multicolumn{6}{c||}{train size: 200} & \multicolumn{6}{c|}{train size: 400} \\ \cline{2-13}
        \multicolumn{1}{c|}{} & \multicolumn{3}{c|}{\textbf{web\_lg}} & \multicolumn{3}{c||}{\textbf{sci\_lg}} & \multicolumn{3}{c|}{\textbf{web\_lg}} & \multicolumn{3}{c|}{\textbf{sci\_lg}} \\
        \hline
        \emph{Classes} (\#) & \emph{P} & \emph{R} & \emph{F1} & \emph{P} & \emph{R} & \emph{F1} & \emph{P} & \emph{R} & \emph{F1} & \emph{P} & \emph{R} & \emph{F1} \\
        \hline
        Observation (185) & 0.81 & 0.83 & 0.82 & 0.85 & 0.94 & 0.89 & 0.81 & 0.82 & 0.81 & 0.84 & 0.93 & 0.88  \\
        \hline
        Input (35) & 0.32 & 0.37 & 0.34 & 0.48 & 0.40 & 0.44 & 0.33 & 0.40 & 0.36 & 0.52 & 0.43 & 0.47 \\
        \hline
        Example (10) & 0.80 & 0.40 & 0.53 & 0.67 & 0.40 & 0.50  & 0.60 & 0.30 & 0.40 &0.50 & 0.30 & 0.37 \\
        \hline
        Other (23) & 0.06 & 0.04 & 0.05 & 0.57 & 0.35 & 0.43 & 0.05 & 0.04 & 0.05 &0.36 & 0.22 & 0.27 \\
        \noalign{\hrule height 2pt}
        \emph{All} (253) & 0.50 & 0.41 & 0.44 & 0.64 & 0.52 & \textbf{0.56} & 0.45 & 0.39 & \textbf{0.41} & 0.55 & 0.47 & 0.50 \\
        \hline
    \end{tabular}
\end{table}

\begin{table}[t]
    \centering
    \caption{\label{tab:resume} Summary of F1 score for different data augmentation approaches. \emph{No Aug.} indicates no Data Augmentation technique was applied, \emph{R/C} indicates row and column inversion technique and \emph{All} indicates row and column inversion technique and random removal of nodes and arcs. }
    \begin{tabular}{|c||c||c|c||c|c|}
        \cline{2-6}
        \multicolumn{1}{c|}{} & \multicolumn{1}{c||}{\textbf{No Aug.}} & \multicolumn{2}{c||}{\textbf{R/C}} & \multicolumn{2}{c|}{\textbf{All}} \\
        \hline
        train size & 64 & 200 & 400 & 200 & 400 \\
        \hline
        \verb|en_core_web_lg| & 0.43 & \textbf{0.45} & 0.44 & \textbf{0.49} & 0.41 \\
        \hline
        \verb|en_core_sci_lg| & 0.52 & \textbf{0.53} & 0.52 & \textbf{0.56} & 0.51 \\
        \hline
    \end{tabular}
    
\end{table}

The main experiments performed are summarized in Table\ \ref{tab:allRes} that compares results obtained  by applying different Data Augmentation techniques and spaCy models \verb|en_core_web_lg|
and \verb|en_core_sci_lg| with baseline results. 
In bold we highlight the most significant results of the F1 score for each technique applied. These values are also summarized in  Table\ \ref{tab:resume} to discuss the outcomes of the experiment. Table \ref{tab:resume} summarizes the best F1 score values obtained considering some DA combinations. We can observe that the models appear rather inaccurate. This is mainly caused by the dataset itself, that is unbalanced toward the Observation class and small in size.
It can also be seen that models using the \verb|en_core_sci_lg| embedding show better results  than those using \verb|en_core_web_lg|, since  the first one is a biomedical-based embedding that is most likely capable of appropriately characterizing and recognizing terms present in the tables extracted from scientific documents.
In particular, models that exploit \verb|en_core_web_lg| and do not use data augmentation techniques turn out to be less accurate and fail to recognize any table of class Other. 
In general, models that employ data augmentation result in higher F1 score values.
Furthermore, observing Table \ref{tab:resume}, it can be seen that better values are obtained for the models in which data augmentation techniques are applied: particularly among these, the one obtained by alternating the inversion of rows and columns with random removal of nodes and arcs is preferable.
\section{Conclusions}
\label{sec:concl}

In this work we presented a GNN model for classifying tables in scientific articles  applying Data Augmentation techniques. The results achieved are promising but still show limitations.
In particular, the imbalance in the available data and a very low number of examples demonstrate that it is difficult to achieve good generalization. However, the use of Data Augmentation techniques made it possible to improve the results obtained by an increase in the F1-Score measure in the ablation studies presented. 
The proposed solution has some aspects that could be deepened or improved to further develop the work started. First, it might be useful to test the implemented Data Augmentation techniques on other datasets to analyze their potential and efficiency. In addition, other Data Augmentation techniques could be implemented, such as adding or removing rows and columns to a table, or improving the techniques already implemented. For example, the row and column recognition techniques could be improved, especially in the case of multi-row tables. 
We conclude by noting how such Data Augmentation techniques applied directly to graphs could prove to be an interesting clue for the application of Graph Neural Networks in the presence of resource-limited datasets, a very common situation in many application domains.
%
%
%
%
%
 \bibliographystyle{splncs04}
 \bibliography{main.bib}
\end{document}